\documentclass[sigconf, nonacm]{acmart}
\usepackage{graphicx}
\AtBeginDocument{%
  }
\usepackage{balance} 

\begin{document}

\title{In-field Calibration of Low-Cost Sensors through XGBoost $\&$ Aggregate Sensor Data}

\author{Kevin Yin}
\affiliation{%
  \institution{University of Michigan}
  \city{Ann Arbor}
  \state{MI}
  \country{USA}}
\email{kevinyin@umich.edu}

\author{Julia Gersey}
\affiliation{%
  \institution{University of Michigan}
  \city{Ann Arbor}
  \state{MI}
  \country{USA}}
\email{gersey@umich.edu}

\author{Pei Zhang}
\affiliation{%
  \institution{University of Michigan}
  \city{Ann Arbor}
  \state{MI}
  \country{USA}}
\email{peizhang@umich.edu}

\renewcommand{\shortauthors}{Yin et al.}

\begin{abstract}
Effective large-scale air quality monitoring necessitates distributed sensing due to the pervasive and harmful nature of particulate matter (PM), particularly in urban environments. However, precision comes at a cost: highly accurate sensors are expensive, limiting the spatial deployments and thus their coverage. As a result, low-cost sensors have become popular, though they are prone to drift caused by environmental sensitivity and manufacturing variability. This paper presents a model for in-field sensor calibration using XGBoost ensemble learning to consolidate data from neighboring sensors. This approach reduces dependence on the presumed accuracy of individual sensors and improves generalization across different locations.
\end{abstract}

\maketitle

\section{Introduction}

Real-time monitoring of air quality is crucial, given the harmful effects of pollution and particulate matter (PM) on human health
\cite{who_air_pollution}. In a 2019 study, the global burden of disease study found that around 12\% of total global deaths can be attributed to air pollution, with PM 2.5 being responsible for 4 to 10 million deaths per year \cite{lelieveld_air_2023}. PM 2.5 has an established effect on human respiratory health, being well documented as a leading cause of respiratory problems, inflammation, lung failure, use of medication and care, and higher mortality rates \cite{anderson_clearing_2012}. Specifically, PM 2.5 is of notable health concern due to it's reduced sizing, allowing for easier infiltration into human respiratory systems as well as into the bloodstream \cite{epa_pm_basics, epa_pm_health_effects}. In addition to its detrimental effect on human health, high quantities of PM can also cause ecological harm, where deposits of fine and course particulates can cause numerous ecological issues such as causing chemical imbalance in soil and rendering ecosystems inhabitable for wildlife and plant life\cite{grantz_ecological_2003}. 

In light of these detrimental effects, efforts to monitor air quality are on the rise. The golden standard for these measurement is through using high precision devices deployed in monitoring stations. These high quality sensors, however, are low in quantity and thus result in sparse data resolution for studies, which can result in failure to differentiate measurements across areas covered by the same monitor and thus exposure misclassifications \cite{datta_statistical_2020}. As a solution, low-cost sensors provide an increasingly popular method for complementing sparse station readings with data points from locations in between or as a fine-grained approach.

\section{Related Work}
Air quality monitoring has been done through both fixed and mobile deployments to monitor a variety of environmental factors \cite{munera2021iot, barot2020air, yin2025survey, gersey2025surveycitywidehomelessnessdetection}. Fine-grained static deployments have leveraged a variety of low-cost sensors to monitor overall air quality and various specific pollution levels \cite{moore_air_2012, cheng_aircloud_2014, jiao_community_2016, tsujita_gas_2005, moltchanov_feasibility_2015, lagerspetz_megasense_2019, gersey_pilot_2023, krupp_towards_2023}. However, these sensor networks face the costs of labor, installation, maintenance, and connectivity. Additionally, low-cost sensors suffer from manufacturing issues and environmental sensitivity that can result in low quality data readings \cite{van_poppel_senseurcity_2023}. To mitigate these limitations, recent projects have leveraged historical data, environmental points of interest, and a variety of machine learning models to estimate and predict pollution levels in unmonitored areas \cite{cheng_maptransfer_2020, liu2024mobiair, chen2018pga, mendez2023machine}.

Mobile air quality sensing has emerged as a complementary approach, increasing spatial coverage by using drones, vehicles, and wearable sensors to measure air pollution across diverse urban environments \cite{apte_high-resolution_2017, noor2024fusion, anjomshoaa_city_2018, badii_real-time_2020, banach_new_2020, dutta_common_2009, nikzad_citisense_2012, maag_w-air_2018, hasenfratz_participatory_2012, gao_mosaic_2016, shirai_toward_2016, gersey2025sniffing}. However, mobile sensing also faces challenges such as temporal sparsity, environmental variability, and calibration drift. Prior work has explored how vehicle speed and driving behavior affect sensor accuracy \cite{liu_delay_2017, liu_individualized_2017, xu_gotcha_2016}, while others have studied strategies for improving sampling by actuating vehicles or incentivizing routes that better cover underrepresented areas \cite{lin_neural-based_2022, joe-wong_taxi-for-all_2021, xu_incentivizing_2019, xu_vehicle_2019, chen_asc_2019}, and optimizing sampling distributions across a region \cite{wu_generative_2020, lin_neural-based_2022}. Even with these adaptive strategies, certain regions may remain undersampled; to address this, machine learning and physics-informed models have been developed to extrapolate pollution data using nearby samples and environmental features \cite{ma2019deep, ma_fine-grained_2020, ma_enhancing_2020, huang2019understanding, chadalavada2024application}.

One of the most persistent challenges in both static and mobile deployments is the poor data quality of low-cost sensors, leading to significant interest in calibration techniques. Calibration may be performed before or after deployment, with pre-deployment calibration relying on access to high-quality reference data to correct for manufacturing variations, while post-deployment calibration addresses environmental conditions and sensor drift \cite{maag_survey_2018}. Both approaches typically involve developing calibration models that dynamically adapt to each deployment. Linear regression remains a simple and popular method for capturing relationships between target and reference variables \cite{villanueva_smart_2023}, but struggles with nonlinear relationships. As a result, a range of machine learning methods—including random forest, gradient boosting, extra trees, BERT, ANNs, and others—have been applied with moderate success \cite{narayana_sens-bert_2023, villanueva_smart_2023}. Calibration models generally fall into monosensor and multisensor categories \cite{villanueva_smart_2023}, with monosensor models training on target and reference sensor pairings over time, and multisensor models leveraging cross-modal readings from sensors with related modalities, such as optical particle counters and electrochemical sensors \cite{van_poppel_senseurcity_2023}.

The effectiveness of both calibration types is often contingent on environmental variables, such as temperature and relative humidity, which strongly influence sensor outputs \cite{kang_calibration_2024}. Furthermore, these models typically rely on co-located reference data, limiting their transferability to new locations without extensive recalibration. Given the increasing use of low-cost sensors in mobile and non-stationary deployments \cite{apte_high-resolution_2017, cai_validation_2014}, scalable spatial calibration models that reduce location dependence are increasingly important. Such models can reduce regional bias and simplify the integration of additional sensors into distributed networks.

Several works have explored spatial calibration models with encouraging results. Chu et al.\ developed a spatial regression mapping model using collocated sensor kernels, reducing RMSE from 17.7–10.5 to 4.1 $\mu$g/m$^3$ \cite{chu_spatial_2020}. GAM networks have been employed to model spatiotemporal variation, yielding RMSE values as low as 4.88 at select locations \cite{lee_efficient_2019}. XGBoost has also proven effective for sensor calibration, achieving RMSE values as low as 4.19 even without spatial features \cite{si_evaluation_2020}. Song et al.\ further demonstrated spatial PM$_{2.5}$ prediction via hyperparameter-optimized XGBoost models, reporting RMSE ranging from 4.86 to 18.64 \cite{song_spatial_2023}. Fast-transfer calibration methods have also been proposed, such as Sens-BERT, which uses BERT-derived embeddings to enable calibration with minimal paired data \cite{narayana_sens-bert_2023}.

In this paper, we propose a method for creating a spatial calibration model of low-cost sensors utilizing XGBoost. Our approach leverages location data from neighboring deployments, along with temperature and relative humidity at each site, to predict calibration adjustments across the sensor network.
  
\section{Data and Methods}
We detail our approach to creating a generalizable, spatially aware calibration model through use of XGBoost. XGBoost was chosen as our model and method of choice due to it's highly effective performance in a number of regression tasks, including non-spatial sensor calibration for other types of sensors and even spatial mapping of air quality values \cite{si_evaluation_2020, tunca_calibrating_2023, song_spatial_2023}. The methodology described can be split into 3 main sections: data acquisition, data processing, and model creation and training.

\subsection{Data Acquisition} 
The dataset utilized in this project is the publicly available SenEURCity dataset collected from 3 European cities (Antwerp, Oslo, Zagreb) using multiple low-cost sensor deployments \cite{van_poppel_senseurcity_2023}. The dataset contains CSV files containing sensor readings from each deployment location, with readings taken from 85 sensors measuring gaseous compound concentrations (NO, CO, CO2, etc) as well as particulate matter concentrations (PM2.5, PM10, PM1) over the span of a year. The datasets also contain reference data readings taken from high accuracy sensing stations. This dataset was chosen for a few key reasons. Firstly, the dataset contains a multitude of sensor deployments in each city as well as location, environmental, and various sensor readings from each location, giving ample data for a model to learn relationships between spatial and environmental variables and calibration amounts. In addition, the dataset contains various different types of sensors deployed at each location, giving enough data to test generalization of the spatial calibration model for other types of sensors beside PM as well as other size PM sensors, for which calibration model quality can often differ \cite{aix_calibration_2023}. 

\subsection{Data Preprocessing}
In order to train a useful model, a couple of pre-processing steps were taken. Firstly, in order to deal with missing data, a number of options were explored, including dropping rows with missing values, imputing with column mean, imputing with median, and forward/backward filling. Of these, the highest RMSE was found to result from forward and backward filling, better capturing more realistic gradual change of conditions and readings over time. Additionally, the number of readings taken by each sensor station was not standardized, resulting in additional time data points with no corresponding input variables to calibrate from. To combat this, data points beyond the lowest number of readings from a single station were dropped and not used in the calibration model. This, however, only resulted in about 10$\%$ of data being dropped from the largest file, resulting in a low impact on the final model performance. Because XGBoost is an ensemble tree method, data normalization has minimal impact on the final learner and thus was not undertaken. Finally, the data was parsed for relevant columns, excluding other types of sensors to reduce learner reliance on co-deployed sensors that are not always guaranteed to be present or accurate in a target location. In the end, 8 input variables are taken from each low-cost sensor location: OPCN3PM25(Alphasense PM2.5 counter), Ref.PM2.5(reference PM2.5 measurement), longitude, latitude, SHT31TI(internal temperature), SHT31TE(external temperature), SHT31HI(internal humidity), and SHT31HE(external humidity). PM2.5 sensor reading, humidity, and temperature are taken as they are proven to be the most impactful variables for a calibration model \cite{kang_calibration_2024}. Reference reading is taken to provide a baseline for training. Longitude and latitude are taken for the model to learn spatial relationships, improving ability of the model to generalize to new sensor deployments in locations not previously seen. For each of the three locations, the data was split into train, validation, and test sets at a ratio of 7:1.5:1.5. 

\subsection{Model Creation $\&$ Training}

XGBoost has shown significant success in creating calibration models for low-cost sensors in previous works, but to the extent of our knowledge haven't yet been used for generating spatially and environmentally aware models predicting calibration amounts. We utilize the XGBoost package in Python to develop a model that learns spatial representations of sensor calibrations that interact with environmental conditions, utilizing the ensemble method's built in hyperparameter turning to develop a robust calibration model. XGBoost offers three types of boosters: Gbtree, Gblinear, and Dart. We tested implementations of the three and found gbtree to provide the best results, returning a root mean squared error (RMSE) about 1 ug/m$^{3}$ less than the gblinear method on the test set and being comparable but simple than Dart. For training, we tested multiple hyperparameter configurations, mainly testing different configurations of learning rate, max tree depth, number of estimators, subsample, colsample bytree, and minimum child weight, as these are most popular hyperparameters to tune and often have the largest impact on the accuracy of the final model \cite{tarwidi_optimized_2023}. Through grid search hyperparameter tuning, we found that the default values were optimal for all values except learning rate and n$\_$estimators, which are best set at 0.16 and 500 respectively. The model was trained for 1000 boosting rounds on a 12th Gen Intel(R) Core(TM) i7-1265U CPU on readings from 34 sensors and over 480000 data points. The learner utilized the held out validation set to guide model learning.

\section{Results}
Our XGBoost model results in satisfactory calibration over an entire sensor network. Results were evaluated in 3 ways: RMSE on the held out test set, ability to generalize to new sensor networks, and ability to generalize to new sensor deployments within the training network. 

The original model was trained on the Antwerp dataset to learn spatial relations and environmental relations to predict calibration values - training and validation loss over the training process are shown in figure 1. When tested on the held out test set, this resulted in an RMSE value of 5.248 over all sensor deployments, resulting in very accurate predictions of the ground truth calibration values. When trained on a reduced number of sensor deployments, the RMSE for the test set was worse when predicting calibration of sensors outside of the spatial area of the training locations. We find that as the amount of deployments considered increases, the model's ability to generalize to new locations improves as well, a conclusion that makes sense given the models reliance on learning spatial relationships. It is also worth noting that the model predicts sensor calibration values for all deployments of sensors in the input kernel, thus the RMSE is a summation of individual sensor errors. In reality, individual sensor calibration values are generally predicted within a margin of 0 to 0.5, showcasing very good performance on the held-out training set for the training location sensor network.

\begin{figure}
\caption{Train and validation RMSE during training}
\includegraphics[scale=.25]{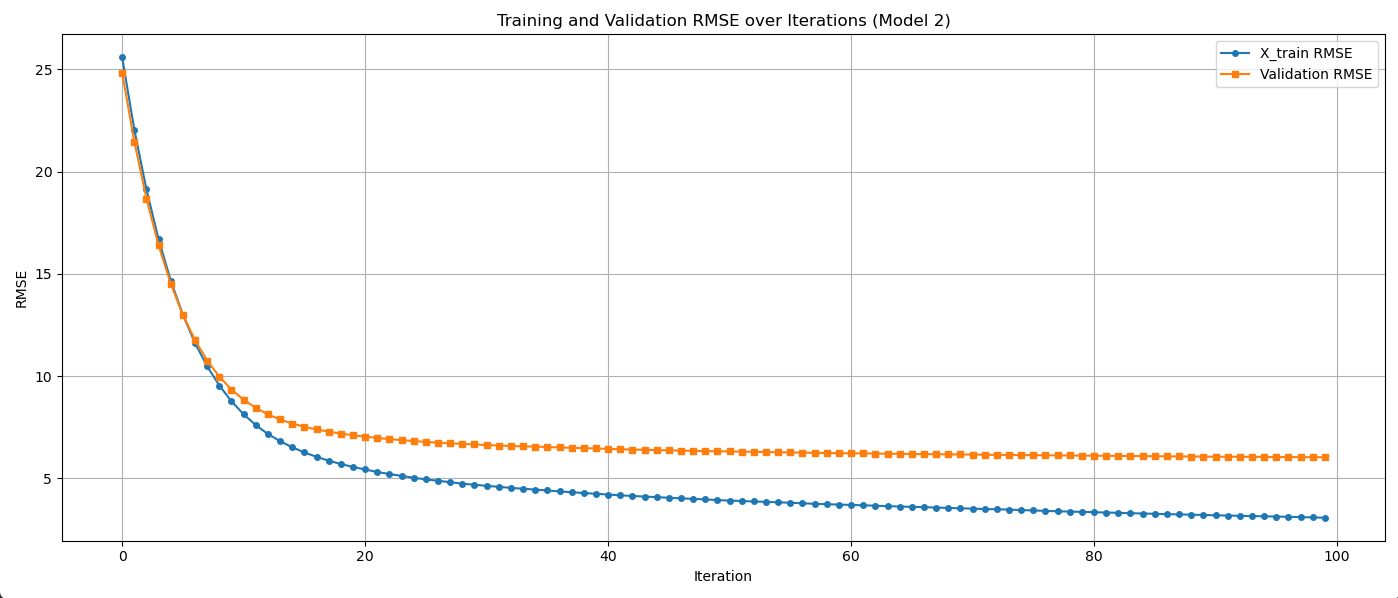}

\end{figure}

When immediately tested on other sensor network locations such as Oslo and Zagreb, RMSE was high, measuring 250.69 on the Oslo dataset and performing similarly poorly on Zagreb. This is to be expected, as the model has not encountered the spatial data of these locations and thus fails to generalize and predict correctly, given that XGBoost is using the gbtree model. Notably, the performance of the gblinear model on a similar task is much better, resulting in an RMSE of 33.35. Though this is a step up from the tree model, it still results in values far off from the ground truth as well as performs worse on the validation set of the training location sensor network - thus we still prefer to explore the tree XGBoost model. We find that with minimal fine turning for only 100 boosting rounds on the tree model, calibration model performance on the collocations can be improved to a lowest RMSE of 6.52, with additional fine-tuning resulting in marginally improved RMSE as seen in figure 2. This improvement gives confidence to the ability of the XGBoost model to learn spatial embeddings from the original model and then easily be fine tuned to new locational data. 

\begin{figure}
\caption{Train and validation RMSE during fine tuning for new array deployment}
\includegraphics[scale=.25]{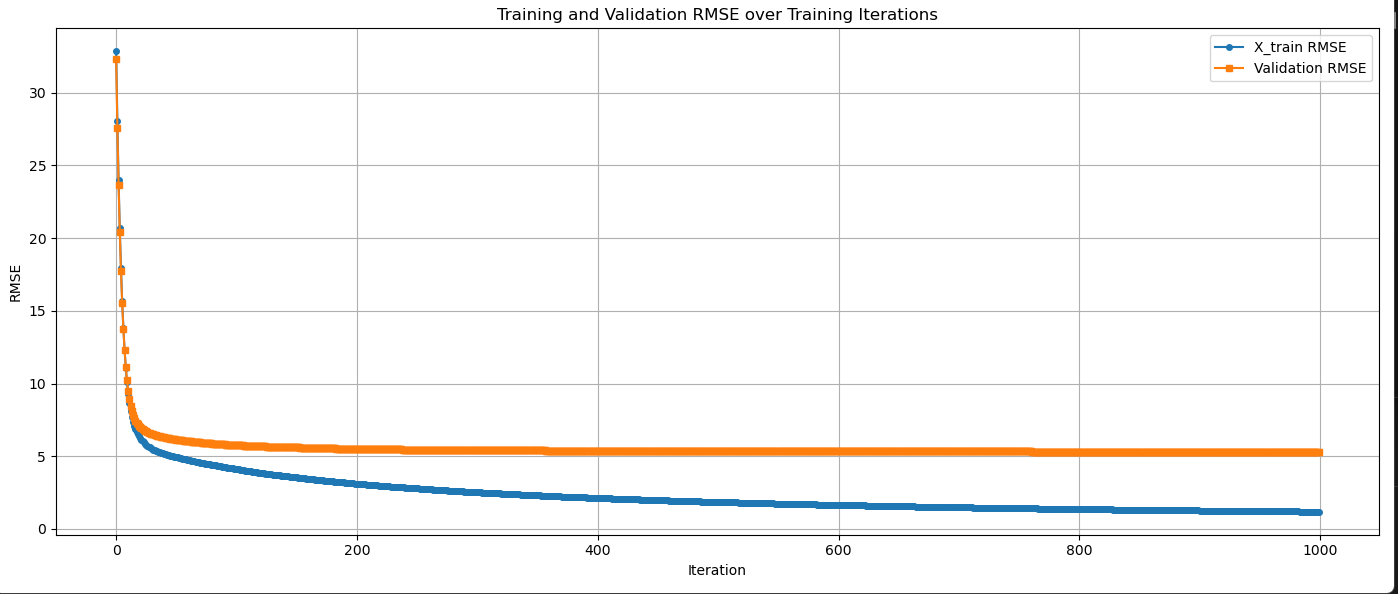}

\end{figure}

With this ability to generalize to completely new sensor network deployments in mind, we then analyze the ability of the network to generalize to a single new sensor added to a location for which the model has already been fine tuned. This is perhaps the most relevant metric given the growing popularity of IoT technology and low-cost sensors in mobile deployments. The ability of the model to generalize to and calibrate new sensors within the context of a spatially aware calibration model is promising as a real-time, accurate, and simple method for calibrating an increasing number of low-cost sensors. Additionally, as experimentally concluded through testing, more data deployment locations also increases the model's generalization ability; thus calibration of additional sensors within a singular network is perhaps the most important use case of this model for measuring performance. When adding a two data points without any fine tuning, RMSE is moderately poor, raising to a max of 24.87 with majority of error correlating to the newly added points. However, minimal fine tuning for about 100 boosting rounds improves performance up to 6.41 RMSE, showing comparable results to the the validation error of the original trained model, as shown in figure 3.
\begin{figure}[b]
\caption{Train and validation RMSE during fine tuning for two sensors}
\includegraphics[scale=.25]{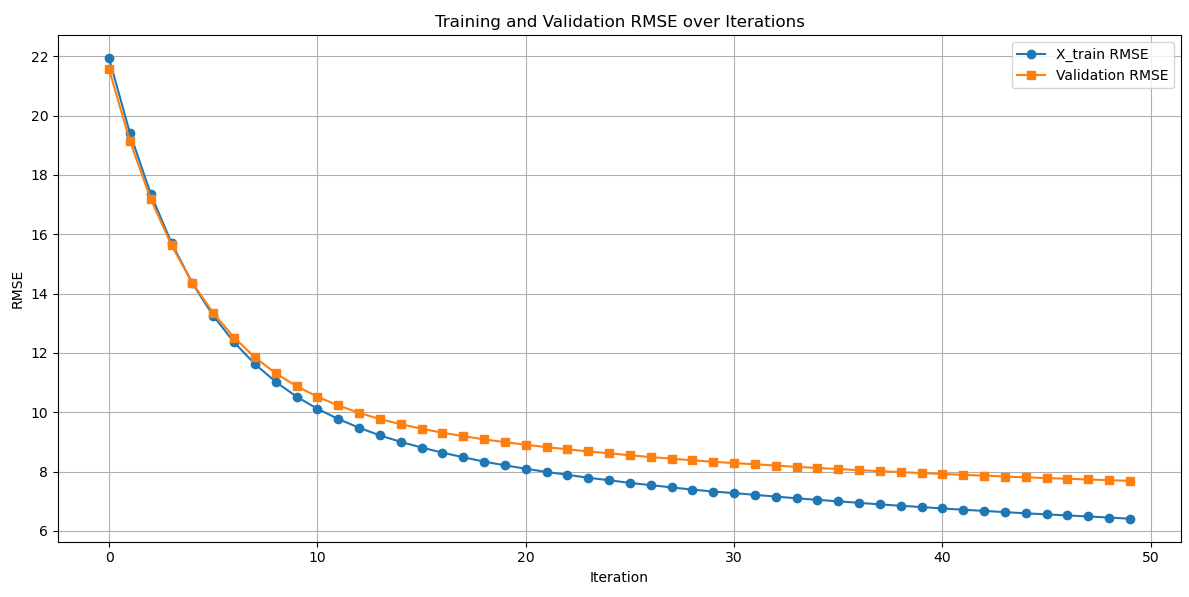}

\end{figure}

\section{Discussion}

Previous papers discussing the use of the XGBoost algorithm for learning nonlinear calibration models focus strictly on a single location, using readings taken from only the target location to calibrate the target \cite{si_evaluation_2020}. Additionally, spatially aware models utilizing XGBoost have previously been made as air quality predictors instead of as sensor calibration models, lacking sensor readings and environmental conditions as inputs \cite{song_spatial_2023}. Our model combines the strictly spatially aware XGBoost mapping model detailed in previous works with sensor calibration models utilizing environmental readings, resulting in an accurate, spatially aware calibration model. Unlike previous works, this model utilizes both environmental and spatial inputs from a multitude of adjacent sensor deployments to calibrate sensors, returning a calibration value for all sensors within the network. Our results show that the XGBoost model has great performance for generalization to new deployments within the spatial area already seen by the model as well as is easy to fine tune to deployment locations completely outside the before seen area. This high level of performance afforded by XGBoost is consistent with previous studies done on the ability of XGBoost to capture spatial relations \cite{song_spatial_2023} as well as in sensor calibration tasks utilizing environmental readings \cite{si_evaluation_2020}. 

As with other works regarding spatial mapping models, future works utilizing altitude as a locational data input could be explored given a dataset containing altitude as a recorded value. Additionally, exploration of different types of NN applied within a spatial and environmental context could potentially create models that outperform XGBoost for generalization tasks. Additionally, this work mainly focused on sensing PM 2.5, given it's danger to human health and resultant need for monitoring. This model of sensor calibration dependent on adjacent deployments and environmental conditions is not sensor specific as it doesn't rely on learning from similar types of sensors, meaning PM 2.5 is not the only type of sensor that this method of calibration could theoretically cover - given the dataset contains multiple types of sensor readings, works examining the effectiveness of the model in calibrating other PM sizes as well as gaseous sensors could be explored. Thus, this calibration method can extend to gaseous compounds or even other types of sensors outside the scope of pollution monitoring such as airborne disease monitoring. Confirmation of the model's ability to generalize to other sensing modalities could be an interesting point of exploration given relevant data.

\section{Conclusions}
Particulate matter is a growing concern for human health and environmental health, with growing rates of pollution only exacerbating the issue. low-cost sensors are a great alternative to expensive air quality monitoring stations, being cheap to deploy and very mobile as well. Though they suffer from low data quality, calibration methods are well documented and utilized for improving the usability of these sensors. The ability to easily generalize an already learned model to new sensor deployments is crucial for the future growth of low-cost sensor deployments. Our proposed XGBoost calibration model is able to generalize to new sensor deployments within the scope of the learned area with moderate success (RMSE = 6.41) and requires minimal fine tuning to new locations outside the learned area (RMSE = 6.52). XGBoost is quick as well as accurate, reducing the burden of data collection and tuning to make calibration of new locations much easier.   

\section{Code}
Our code is publicly available at the following GitHub repository url: \url{https://github.com/virident/sensor_cal}

\balance
\bibliographystyle{ACM-Reference-Format}
\bibliography{main.bib}


\begin{thebibliography}{62}


\ifx \showCODEN    \undefined \def \showCODEN     #1{\unskip}     \fi
\ifx \showISBNx    \undefined \def \showISBNx     #1{\unskip}     \fi
\ifx \showISBNxiii \undefined \def \showISBNxiii  #1{\unskip}     \fi
\ifx \showISSN     \undefined \def \showISSN      #1{\unskip}     \fi
\ifx \showLCCN     \undefined \def \showLCCN      #1{\unskip}     \fi
\ifx \shownote     \undefined \def \shownote      #1{#1}          \fi
\ifx \showarticletitle \undefined \def \showarticletitle #1{#1}   \fi
\ifx \showURL      \undefined \def \showURL       {\relax}        \fi
\providecommand\bibfield[2]{#2}
\providecommand\bibinfo[2]{#2}
\providecommand\natexlab[1]{#1}
\providecommand\showeprint[2][]{arXiv:#2}

\bibitem[Aix et~al\mbox{.}(2023)]%
        {aix_calibration_2023}
\bibfield{author}{\bibinfo{person}{Marie-Laure Aix}, \bibinfo{person}{Seán Schmitz}, {and} \bibinfo{person}{Dominique~J. Bicout}.} \bibinfo{year}{2023}\natexlab{}.
\newblock \showarticletitle{Calibration methodology of low-cost sensors for high-quality monitoring of fine particulate matter}.
\newblock   \bibinfo{volume}{889} (\bibinfo{year}{2023}), \bibinfo{pages}{164063}.
\newblock
\showISSN{0048-9697}
\href{https://doi.org/10.1016/j.scitotenv.2023.164063}{doi:\nolinkurl{10.1016/j.scitotenv.2023.164063}}


\bibitem[Anderson et~al\mbox{.}(2012)]%
        {anderson_clearing_2012}
\bibfield{author}{\bibinfo{person}{Jonathan~O. Anderson}, \bibinfo{person}{Josef~G. Thundiyil}, {and} \bibinfo{person}{Andrew Stolbach}.} \bibinfo{year}{2012}\natexlab{}.
\newblock \showarticletitle{Clearing the Air: A Review of the Effects of Particulate Matter Air Pollution on Human Health}.
\newblock  \bibinfo{volume}{8}, \bibinfo{number}{2} (\bibinfo{year}{2012}), \bibinfo{pages}{166--175}.
\newblock
\showISSN{1937-6995}
\href{https://doi.org/10.1007/s13181-011-0203-1}{doi:\nolinkurl{10.1007/s13181-011-0203-1}}


\bibitem[Anjomshoaa et~al\mbox{.}(2018)]%
        {anjomshoaa_city_2018}
\bibfield{author}{\bibinfo{person}{Amin Anjomshoaa}, \bibinfo{person}{Fábio Duarte}, \bibinfo{person}{Daniël Rennings}, \bibinfo{person}{Thomas~J. Matarazzo}, \bibinfo{person}{Priyanka deSouza}, {and} \bibinfo{person}{Carlo Ratti}.} \bibinfo{year}{2018}\natexlab{}.
\newblock \showarticletitle{City {Scanner}: {Building} and {Scheduling} a {Mobile} {Sensing} {Platform} for {Smart} {City} {Services}}.
\newblock \bibinfo{journal}{\emph{IEEE Internet of Things Journal}} \bibinfo{volume}{5}, \bibinfo{number}{6} (\bibinfo{date}{Dec.} \bibinfo{year}{2018}), \bibinfo{pages}{4567--4579}.
\newblock
\showISSN{2327-4662}
\href{https://doi.org/10.1109/JIOT.2018.2839058}{doi:\nolinkurl{10.1109/JIOT.2018.2839058}}


\bibitem[Apte et~al\mbox{.}(2017)]%
        {apte_high-resolution_2017}
\bibfield{author}{\bibinfo{person}{Joshua~S. Apte}, \bibinfo{person}{Kyle~P. Messier}, \bibinfo{person}{Shahzad Gani}, \bibinfo{person}{Michael Brauer}, \bibinfo{person}{Thomas~W. Kirchstetter}, \bibinfo{person}{Melissa~M. Lunden}, \bibinfo{person}{Julian~D. Marshall}, \bibinfo{person}{Christopher~J. Portier}, \bibinfo{person}{Roel~C.H. Vermeulen}, {and} \bibinfo{person}{Steven~P. Hamburg}.} \bibinfo{year}{2017}\natexlab{}.
\newblock \showarticletitle{High-Resolution Air Pollution Mapping with Google Street View Cars: Exploiting Big Data}.
\newblock  \bibinfo{volume}{51}, \bibinfo{number}{12} (\bibinfo{year}{2017}), \bibinfo{pages}{6999--7008}.
\newblock
\showISSN{0013-936X, 1520-5851}
\href{https://doi.org/10.1021/acs.est.7b00891}{doi:\nolinkurl{10.1021/acs.est.7b00891}}


\bibitem[Badii et~al\mbox{.}(2020)]%
        {badii_real-time_2020}
\bibfield{author}{\bibinfo{person}{Claudio Badii}, \bibinfo{person}{Stefano Bilotta}, \bibinfo{person}{Daniele Cenni}, \bibinfo{person}{Angelo Difino}, \bibinfo{person}{Paolo Nesi}, \bibinfo{person}{Irene Paoli}, {and} \bibinfo{person}{Michela Paolucci}.} \bibinfo{year}{2020}\natexlab{}.
\newblock \showarticletitle{Real-{Time} {Automatic} {Air} {Pollution} {Services} from {IOT} {Data} {Network}}. In \bibinfo{booktitle}{\emph{2020 {IEEE} {Symposium} on {Computers} and {Communications} ({ISCC})}}. \bibinfo{pages}{1--6}.
\newblock
\href{https://doi.org/10.1109/ISCC50000.2020.9219580}{doi:\nolinkurl{10.1109/ISCC50000.2020.9219580}}
\newblock
\shownote{ISSN: 2642-7389}.


\bibitem[Banach et~al\mbox{.}(2020)]%
        {banach_new_2020}
\bibfield{author}{\bibinfo{person}{Marzena Banach}, \bibinfo{person}{Tomasz Talaśka}, \bibinfo{person}{Jakub Dalecki}, {and} \bibinfo{person}{Rafał Długosz}.} \bibinfo{year}{2020}\natexlab{}.
\newblock \showarticletitle{New technologies for smart cities – high-resolution air pollution maps based on intelligent sensors}.
\newblock \bibinfo{journal}{\emph{Concurrency and Computation: Practice and Experience}} \bibinfo{volume}{32}, \bibinfo{number}{13} (\bibinfo{year}{2020}), \bibinfo{pages}{e5179}.
\newblock
\showISSN{1532-0634}
\href{https://doi.org/10.1002/cpe.5179}{doi:\nolinkurl{10.1002/cpe.5179}}
\newblock
\shownote{\_eprint: https://onlinelibrary.wiley.com/doi/pdf/10.1002/cpe.5179}.


\bibitem[Barot and Kapadia(2020)]%
        {barot2020air}
\bibfield{author}{\bibinfo{person}{Virendra Barot} {and} \bibinfo{person}{Viral Kapadia}.} \bibinfo{year}{2020}\natexlab{}.
\newblock \showarticletitle{Air quality monitoring systems using IoT: a review}. In \bibinfo{booktitle}{\emph{2020 international conference on computational performance evaluation (ComPE)}}. IEEE, \bibinfo{pages}{226--231}.
\newblock


\bibitem[Cai et~al\mbox{.}(2014)]%
        {cai_validation_2014}
\bibfield{author}{\bibinfo{person}{Jing Cai}, \bibinfo{person}{Beizhan Yan}, \bibinfo{person}{James Ross}, \bibinfo{person}{Danian Zhang}, \bibinfo{person}{Patrick~L. Kinney}, \bibinfo{person}{Matthew~S. Perzanowski}, \bibinfo{person}{{KyungHwa} Jung}, \bibinfo{person}{Rachel Miller}, {and} \bibinfo{person}{Steven~N. Chillrud}.} \bibinfo{year}{2014}\natexlab{}.
\newblock \showarticletitle{Validation of {MicroAeth}® as a Black Carbon Monitor for Fixed-Site Measurement and Optimization for Personal Exposure Characterization}.
\newblock  \bibinfo{volume}{14}, \bibinfo{number}{1} (\bibinfo{year}{2014}), \bibinfo{pages}{1--9}.
\newblock
\showISSN{2071-1409}
\href{https://doi.org/10.4209/aaqr.2013.03.0088}{doi:\nolinkurl{10.4209/aaqr.2013.03.0088}}


\bibitem[Chadalavada et~al\mbox{.}(2024)]%
        {chadalavada2024application}
\bibfield{author}{\bibinfo{person}{Sreeni Chadalavada}, \bibinfo{person}{Oliver Faust}, \bibinfo{person}{Massimo Salvi}, \bibinfo{person}{Silvia Seoni}, \bibinfo{person}{Nawin Raj}, \bibinfo{person}{U Raghavendra}, \bibinfo{person}{Anjan Gudigar}, \bibinfo{person}{Prabal~Datta Barua}, \bibinfo{person}{Filippo Molinari}, {and} \bibinfo{person}{Rajendra Acharya}.} \bibinfo{year}{2024}\natexlab{}.
\newblock \showarticletitle{Application of artificial intelligence in air pollution monitoring and forecasting: A systematic review}.
\newblock \bibinfo{journal}{\emph{Environmental Modelling \& Software}} (\bibinfo{year}{2024}), \bibinfo{pages}{106312}.
\newblock


\bibitem[Chen et~al\mbox{.}(2019)]%
        {chen_asc_2019}
\bibfield{author}{\bibinfo{person}{Xinlei Chen}, \bibinfo{person}{Susu Xu}, \bibinfo{person}{Haohao Fu}, \bibinfo{person}{Carlee Joe-Wong}, \bibinfo{person}{Lin Zhang}, \bibinfo{person}{Hae~Young Noh}, {and} \bibinfo{person}{Pei Zhang}.} \bibinfo{year}{2019}\natexlab{}.
\newblock \showarticletitle{{ASC}: actuation system for city-wide crowdsensing with ride-sharing vehicular platform}. In \bibinfo{booktitle}{\emph{Proceedings of the {Fourth} {Workshop} on {International} {Science} of {Smart} {City} {Operations} and {Platforms} {Engineering}}}. \bibinfo{publisher}{ACM}, \bibinfo{address}{Montreal Quebec Canada}, \bibinfo{pages}{19--24}.
\newblock
\showISBNx{978-1-4503-6703-5}
\href{https://doi.org/10.1145/3313237.3313299}{doi:\nolinkurl{10.1145/3313237.3313299}}


\bibitem[Chen et~al\mbox{.}(2018)]%
        {chen2018pga}
\bibfield{author}{\bibinfo{person}{Xinlei Chen}, \bibinfo{person}{Xiangxiang Xu}, \bibinfo{person}{Xinyu Liu}, \bibinfo{person}{Shijia Pan}, \bibinfo{person}{Jiayou He}, \bibinfo{person}{Hae~Young Noh}, \bibinfo{person}{Lin Zhang}, {and} \bibinfo{person}{Pei Zhang}.} \bibinfo{year}{2018}\natexlab{}.
\newblock \showarticletitle{Pga: Physics guided and adaptive approach for mobile fine-grained air pollution estimation}. In \bibinfo{booktitle}{\emph{Proceedings of the 2018 ACM International Joint Conference and 2018 International Symposium on Pervasive and Ubiquitous Computing and Wearable Computers}}. \bibinfo{pages}{1321--1330}.
\newblock


\bibitem[Cheng et~al\mbox{.}(2020)]%
        {cheng_maptransfer_2020}
\bibfield{author}{\bibinfo{person}{Yun Cheng}, \bibinfo{person}{Xiaoxi He}, \bibinfo{person}{Zimu Zhou}, {and} \bibinfo{person}{Lothar Thiele}.} \bibinfo{year}{2020}\natexlab{}.
\newblock \showarticletitle{{MapTransfer} : {Urban} {Air} {Quality} {Map} {Generation} for {Downscaled} {Sensor} {Deployments}}. In \bibinfo{booktitle}{\emph{2020 {IEEE}/{ACM} {Fifth} {International} {Conference} on {Internet}-of-{Things} {Design} and {Implementation} ({IoTDI})}}. \bibinfo{pages}{14--26}.
\newblock
\href{https://doi.org/10.1109/IoTDI49375.2020.00010}{doi:\nolinkurl{10.1109/IoTDI49375.2020.00010}}


\bibitem[Cheng et~al\mbox{.}(2014)]%
        {cheng_aircloud_2014}
\bibfield{author}{\bibinfo{person}{Yun Cheng}, \bibinfo{person}{Xiucheng Li}, \bibinfo{person}{Zhijun Li}, \bibinfo{person}{Shouxu Jiang}, \bibinfo{person}{Yilong Li}, \bibinfo{person}{Ji Jia}, {and} \bibinfo{person}{Xiaofan Jiang}.} \bibinfo{year}{2014}\natexlab{}.
\newblock \showarticletitle{{AirCloud}: a cloud-based air-quality monitoring system for everyone}. In \bibinfo{booktitle}{\emph{Proceedings of the 12th {ACM} {Conference} on {Embedded} {Network} {Sensor} {Systems}}}. \bibinfo{publisher}{ACM}, \bibinfo{address}{Memphis Tennessee}, \bibinfo{pages}{251--265}.
\newblock
\showISBNx{978-1-4503-3143-2}
\href{https://doi.org/10.1145/2668332.2668346}{doi:\nolinkurl{10.1145/2668332.2668346}}


\bibitem[Chu et~al\mbox{.}(2020)]%
        {chu_spatial_2020}
\bibfield{author}{\bibinfo{person}{Hone-Jay Chu}, \bibinfo{person}{Muhammad~Zeeshan Ali}, {and} \bibinfo{person}{Yu-Chen He}.} \bibinfo{year}{2020}\natexlab{}.
\newblock \showarticletitle{Spatial calibration and {PM2}.5 mapping of low-cost air quality sensors}.
\newblock \bibinfo{journal}{\emph{Scientific Reports}} \bibinfo{volume}{10}, \bibinfo{number}{1} (\bibinfo{date}{Dec.} \bibinfo{year}{2020}), \bibinfo{pages}{22079}.
\newblock
\showISSN{2045-2322}
\href{https://doi.org/10.1038/s41598-020-79064-w}{doi:\nolinkurl{10.1038/s41598-020-79064-w}}


\bibitem[Datta et~al\mbox{.}(2020)]%
        {datta_statistical_2020}
\bibfield{author}{\bibinfo{person}{Abhirup Datta}, \bibinfo{person}{Arkajyoti Saha}, \bibinfo{person}{Misti~Levy Zamora}, \bibinfo{person}{Colby Buehler}, \bibinfo{person}{Lei Hao}, \bibinfo{person}{Fulizi Xiong}, \bibinfo{person}{Drew~R. Gentner}, {and} \bibinfo{person}{Kirsten Koehler}.} \bibinfo{year}{2020}\natexlab{}.
\newblock \showarticletitle{Statistical field calibration of a low-cost {PM}2.5 monitoring network in Baltimore}.
\newblock   \bibinfo{volume}{242} (\bibinfo{year}{2020}), \bibinfo{pages}{117761}.
\newblock
\showISSN{1352-2310}
\href{https://doi.org/10.1016/j.atmosenv.2020.117761}{doi:\nolinkurl{10.1016/j.atmosenv.2020.117761}}


\bibitem[Dutta et~al\mbox{.}(2009)]%
        {dutta_common_2009}
\bibfield{author}{\bibinfo{person}{Prabal Dutta}, \bibinfo{person}{Paul~M. Aoki}, \bibinfo{person}{Neil Kumar}, \bibinfo{person}{Alan Mainwaring}, \bibinfo{person}{Chris Myers}, \bibinfo{person}{Wesley Willett}, {and} \bibinfo{person}{Allison Woodruff}.} \bibinfo{year}{2009}\natexlab{}.
\newblock \showarticletitle{Common {Sense}: participatory urban sensing using a network of handheld air quality monitors}. In \bibinfo{booktitle}{\emph{Proceedings of the 7th {ACM} {Conference} on {Embedded} {Networked} {Sensor} {Systems}}}. \bibinfo{publisher}{ACM}, \bibinfo{address}{Berkeley California}, \bibinfo{pages}{349--350}.
\newblock
\showISBNx{978-1-60558-519-2}
\href{https://doi.org/10.1145/1644038.1644095}{doi:\nolinkurl{10.1145/1644038.1644095}}


\bibitem[{Environmental Protection Agency}(2022a)]%
        {epa_pm_health_effects}
\bibfield{author}{\bibinfo{person}{{Environmental Protection Agency}}.} \bibinfo{year}{2022}\natexlab{a}.
\newblock \bibinfo{title}{Health and Environmental Effects of Particulate Matter (PM)}.
\newblock \bibinfo{howpublished}{\url{https://www.epa.gov/pm-pollution/health-andenvironmental-effects-particulate-matter-pm}}.
\newblock
\newblock
\shownote{Accessed: 2025-03-29}.


\bibitem[{Environmental Protection Agency}(2022b)]%
        {epa_pm_basics}
\bibfield{author}{\bibinfo{person}{{Environmental Protection Agency}}.} \bibinfo{year}{2022}\natexlab{b}.
\newblock \bibinfo{title}{Particulate Matter (PM) Basics}.
\newblock \bibinfo{howpublished}{\url{https://www.epa.gov/pm-pollution/particulate-matter-pm-basics}}.
\newblock
\newblock
\shownote{Accessed: 2025-03-29}.


\bibitem[Gao et~al\mbox{.}(2016)]%
        {gao_mosaic_2016}
\bibfield{author}{\bibinfo{person}{Yi Gao}, \bibinfo{person}{Wei Dong}, \bibinfo{person}{Kai Guo}, \bibinfo{person}{Xue Liu}, \bibinfo{person}{Yuan Chen}, \bibinfo{person}{Xiaojin Liu}, \bibinfo{person}{Jiajun Bu}, {and} \bibinfo{person}{Chun Chen}.} \bibinfo{year}{2016}\natexlab{}.
\newblock \showarticletitle{Mosaic: {A} low-cost mobile sensing system for urban air quality monitoring}. In \bibinfo{booktitle}{\emph{{IEEE} {INFOCOM} 2016 - {The} 35th {Annual} {IEEE} {International} {Conference} on {Computer} {Communications}}}. \bibinfo{pages}{1--9}.
\newblock
\href{https://doi.org/10.1109/INFOCOM.2016.7524478}{doi:\nolinkurl{10.1109/INFOCOM.2016.7524478}}


\bibitem[Gersey et~al\mbox{.}(2025a)]%
        {gersey2025sniffing}
\bibfield{author}{\bibinfo{person}{Julia Gersey}, \bibinfo{person}{Jatin Aggarwal}, \bibinfo{person}{Jiale Zhang}, \bibinfo{person}{Jesse Codling}, {and} \bibinfo{person}{Pei Zhang}.} \bibinfo{year}{2025}\natexlab{a}.
\newblock \showarticletitle{Sniffing Out the City-Vehicular Multimodal Sensing for Environmental and Infrastructure Analysis}. In \bibinfo{booktitle}{\emph{Proceedings of the 23rd ACM Conference on Embedded Networked Sensor Systems}}. \bibinfo{pages}{632--633}.
\newblock


\bibitem[Gersey et~al\mbox{.}(2025b)]%
        {gersey2025surveycitywidehomelessnessdetection}
\bibfield{author}{\bibinfo{person}{Julia Gersey}, \bibinfo{person}{Rose Allegrette}, \bibinfo{person}{Joshua Lian}, \bibinfo{person}{Zawad Munshi}, {and} \bibinfo{person}{Aarti Phatke}.} \bibinfo{year}{2025}\natexlab{b}.
\newblock \bibinfo{title}{Survey of City-Wide Homelessness Detection Through Environmental Sensing}.
\newblock
\showeprint[arxiv]{2503.11727}~[cs.CY]
\urldef\tempurl%
\url{https://arxiv.org/abs/2503.11727}
\showURL{%
\tempurl}


\bibitem[Gersey et~al\mbox{.}(2023)]%
        {gersey_pilot_2023}
\bibfield{author}{\bibinfo{person}{Julia Gersey}, \bibinfo{person}{Brian Krupp}, {and} \bibinfo{person}{Jonathon Fagert}.} \bibinfo{year}{2023}\natexlab{}.
\newblock \showarticletitle{Pilot {Study} of {Deploying} {IoT} {Micro} {Air} {Quality} {Sensors} in an {Urban} {Environment}: {Lessons} {Learned}}.
\newblock \bibinfo{journal}{\emph{J. Comput. Sci. Coll.}} \bibinfo{volume}{39}, \bibinfo{number}{4} (\bibinfo{date}{Oct.} \bibinfo{year}{2023}), \bibinfo{pages}{74--83}.
\newblock
\showISSN{1937-4771}


\bibitem[Grantz et~al\mbox{.}(2003)]%
        {grantz_ecological_2003}
\bibfield{author}{\bibinfo{person}{D.~A Grantz}, \bibinfo{person}{J.~H.~B Garner}, {and} \bibinfo{person}{D.~W Johnson}.} \bibinfo{year}{2003}\natexlab{}.
\newblock \showarticletitle{Ecological effects of particulate matter}.
\newblock  \bibinfo{volume}{29}, \bibinfo{number}{2} (\bibinfo{year}{2003}), \bibinfo{pages}{213--239}.
\newblock
\showISSN{0160-4120}
\href{https://doi.org/10.1016/S0160-4120(02)00181-2}{doi:\nolinkurl{10.1016/S0160-4120(02)00181-2}}


\bibitem[Hasenfratz et~al\mbox{.}(2012)]%
        {hasenfratz_participatory_2012}
\bibfield{author}{\bibinfo{person}{David Hasenfratz}, \bibinfo{person}{Olga Saukh}, \bibinfo{person}{Silvan Sturzenegger}, {and} \bibinfo{person}{Lothar Thiele}.} \bibinfo{year}{2012}\natexlab{}.
\newblock \bibinfo{title}{Participatory {Air} {Pollution} {Monitoring} {Using} {Smartphones}}.
\newblock
\urldef\tempurl%
\url{https://tik-old.ee.ethz.ch/file//b6c2122d089d2ef88348a74ddf2906dc/HSST2012.pdf}
\showURL{%
\tempurl}


\bibitem[Huang et~al\mbox{.}(2019)]%
        {huang2019understanding}
\bibfield{author}{\bibinfo{person}{Jiayi Huang}, \bibinfo{person}{Ning Liu}, \bibinfo{person}{Rui Ma}, \bibinfo{person}{Xinyu Liu}, \bibinfo{person}{Yue Wang}, {and} \bibinfo{person}{Lin Zhang}.} \bibinfo{year}{2019}\natexlab{}.
\newblock \showarticletitle{Understanding air pollution patterns in city based on minute-level event detection}. In \bibinfo{booktitle}{\emph{Proceedings of the 17th Conference on Embedded Networked Sensor Systems}}. \bibinfo{pages}{448--449}.
\newblock


\bibitem[Jiao et~al\mbox{.}(2016)]%
        {jiao_community_2016}
\bibfield{author}{\bibinfo{person}{Wan Jiao}, \bibinfo{person}{Gayle Hagler}, \bibinfo{person}{Ronald Williams}, \bibinfo{person}{Robert Sharpe}, \bibinfo{person}{Ryan Brown}, \bibinfo{person}{Daniel Garver}, \bibinfo{person}{Robert Judge}, \bibinfo{person}{Motria Caudill}, \bibinfo{person}{Joshua Rickard}, \bibinfo{person}{Michael Davis}, \bibinfo{person}{Lewis Weinstock}, \bibinfo{person}{Susan Zimmer-Dauphinee}, {and} \bibinfo{person}{Ken Buckley}.} \bibinfo{year}{2016}\natexlab{}.
\newblock \showarticletitle{Community {Air} {Sensor} {Network} ({CAIRSENSE}) project: evaluation of low-cost sensor performance in a suburban environment in the southeastern {United} {States}}.
\newblock \bibinfo{journal}{\emph{Atmospheric Measurement Techniques}} \bibinfo{volume}{9}, \bibinfo{number}{11} (\bibinfo{date}{Nov.} \bibinfo{year}{2016}), \bibinfo{pages}{5281--5292}.
\newblock
\showISSN{1867-1381}
\href{https://doi.org/10.5194/amt-9-5281-2016}{doi:\nolinkurl{10.5194/amt-9-5281-2016}}
\newblock
\shownote{Publisher: Copernicus GmbH}.


\bibitem[Joe-Wong et~al\mbox{.}(2021)]%
        {joe-wong_taxi-for-all_2021}
\bibfield{author}{\bibinfo{person}{Carlee Joe-Wong}, \bibinfo{person}{Haeyoung Noh}, \bibinfo{person}{Pei Zhang}, {and} \bibinfo{person}{Carnegie Mellon~University Mobility21}.} \bibinfo{year}{2021}\natexlab{}.
\newblock \bibinfo{booktitle}{\emph{Taxi-for-all: {Incentivized} {Taxi} {Actuation} {System} for {Balanced} {Area}-wide {Service}}}.
\newblock \bibinfo{type}{{T}echnical {R}eport}.
\newblock
\urldef\tempurl%
\url{https://rosap.ntl.bts.gov/view/dot/59923}
\showURL{%
\tempurl}


\bibitem[Kang and Choi(2024)]%
        {kang_calibration_2024}
\bibfield{author}{\bibinfo{person}{Jiwoo Kang} {and} \bibinfo{person}{Kanghyeok Choi}.} \bibinfo{year}{2024}\natexlab{}.
\newblock \showarticletitle{Calibration Methods for Low-Cost Particulate Matter Sensors Considering Seasonal Variability}.
\newblock  \bibinfo{volume}{24}, \bibinfo{number}{10} (\bibinfo{year}{2024}), \bibinfo{pages}{3023}.
\newblock
\showISSN{1424-8220}
\href{https://doi.org/10.3390/s24103023}{doi:\nolinkurl{10.3390/s24103023}}


\bibitem[Krupp et~al\mbox{.}(2023)]%
        {krupp_towards_2023}
\bibfield{author}{\bibinfo{person}{Brian Krupp}, \bibinfo{person}{Julia Gersey}, \bibinfo{person}{Jonathon Fagert}, {and} \bibinfo{person}{Tony Mlady}.} \bibinfo{year}{2023}\natexlab{}.
\newblock \showarticletitle{Towards {Fine}-{Grained} {Air} {Quality} {Sensing} in {Urban} {Environments}}. In \bibinfo{booktitle}{\emph{Proceedings of the 20th {ACM} {Conference} on {Embedded} {Networked} {Sensor} {Systems}}} \emph{(\bibinfo{series}{{SenSys} '22})}. \bibinfo{publisher}{Association for Computing Machinery}, \bibinfo{address}{New York, NY, USA}, \bibinfo{pages}{851--852}.
\newblock
\showISBNx{978-1-4503-9886-2}
\href{https://doi.org/10.1145/3560905.3568071}{doi:\nolinkurl{10.1145/3560905.3568071}}


\bibitem[Lagerspetz et~al\mbox{.}(2019)]%
        {lagerspetz_megasense_2019}
\bibfield{author}{\bibinfo{person}{Eemil Lagerspetz}, \bibinfo{person}{Naser~Hossein Motlagh}, \bibinfo{person}{Martha Arbayani~Zaidan}, \bibinfo{person}{Pak~L. Fung}, \bibinfo{person}{Julien Mineraud}, \bibinfo{person}{Samu Varjonen}, \bibinfo{person}{Matti Siekkinen}, \bibinfo{person}{Petteri Nurmi}, \bibinfo{person}{Yutaka Matsumi}, \bibinfo{person}{Sasu Tarkoma}, {and} \bibinfo{person}{Tareq Hussein}.} \bibinfo{year}{2019}\natexlab{}.
\newblock \showarticletitle{{MegaSense}: {Feasibility} of {Low}-{Cost} {Sensors} for {Pollution} {Hot}-spot {Detection}}. In \bibinfo{booktitle}{\emph{2019 {IEEE} 17th {International} {Conference} on {Industrial} {Informatics} ({INDIN})}}, Vol.~\bibinfo{volume}{1}. \bibinfo{pages}{1083--1090}.
\newblock
\href{https://doi.org/10.1109/INDIN41052.2019.8971963}{doi:\nolinkurl{10.1109/INDIN41052.2019.8971963}}
\newblock
\shownote{ISSN: 2378-363X}.


\bibitem[Lee et~al\mbox{.}(2019)]%
        {lee_efficient_2019}
\bibfield{author}{\bibinfo{person}{Chieh-Han Lee}, \bibinfo{person}{Yeuh-Bin Wang}, {and} \bibinfo{person}{Hwa-Lung Yu}.} \bibinfo{year}{2019}\natexlab{}.
\newblock \showarticletitle{An efficient spatiotemporal data calibration approach for the low-cost {PM}2.5 sensing network: A case study in Taiwan}.
\newblock   \bibinfo{volume}{130} (\bibinfo{year}{2019}), \bibinfo{pages}{104838}.
\newblock
\showISSN{0160-4120}
\href{https://doi.org/10.1016/j.envint.2019.05.032}{doi:\nolinkurl{10.1016/j.envint.2019.05.032}}


\bibitem[Lelieveld et~al\mbox{.}(2023)]%
        {lelieveld_air_2023}
\bibfield{author}{\bibinfo{person}{Jos Lelieveld}, \bibinfo{person}{Andy Haines}, \bibinfo{person}{Richard Burnett}, \bibinfo{person}{Cathryn Tonne}, \bibinfo{person}{Klaus Klingmüller}, \bibinfo{person}{Thomas Münzel}, {and} \bibinfo{person}{Andrea Pozzer}.} \bibinfo{year}{2023}\natexlab{}.
\newblock \showarticletitle{Air pollution deaths attributable to fossil fuels: observational and modelling study}.
\newblock  (\bibinfo{year}{2023}), \bibinfo{pages}{e077784}.
\newblock
\showISSN{1756-1833}
\href{https://doi.org/10.1136/bmj-2023-077784}{doi:\nolinkurl{10.1136/bmj-2023-077784}}


\bibitem[Lin et~al\mbox{.}(2022)]%
        {lin_neural-based_2022}
\bibfield{author}{\bibinfo{person}{Shouxu Lin}, \bibinfo{person}{Yuhang Yao}, \bibinfo{person}{Pei Zhang}, \bibinfo{person}{Hae~Young Noh}, {and} \bibinfo{person}{Carlee Joe-Wong}.} \bibinfo{year}{2022}\natexlab{}.
\newblock \showarticletitle{A neural-based bandit approach to mobile crowdsourcing}. In \bibinfo{booktitle}{\emph{Proceedings of the 23rd {Annual} {International} {Workshop} on {Mobile} {Computing} {Systems} and {Applications}}}. \bibinfo{publisher}{ACM}, \bibinfo{address}{Tempe Arizona}, \bibinfo{pages}{15--21}.
\newblock
\showISBNx{978-1-4503-9218-1}
\href{https://doi.org/10.1145/3508396.3512886}{doi:\nolinkurl{10.1145/3508396.3512886}}


\bibitem[Liu et~al\mbox{.}(2017a)]%
        {liu_delay_2017}
\bibfield{author}{\bibinfo{person}{Xinyu Liu}, \bibinfo{person}{Xinlei Chen}, \bibinfo{person}{Xiangxiang Xu}, \bibinfo{person}{Enhan Mai}, \bibinfo{person}{Hae~Young Noh}, \bibinfo{person}{Pei Zhang}, {and} \bibinfo{person}{Lin Zhang}.} \bibinfo{year}{2017}\natexlab{a}.
\newblock \showarticletitle{Delay {Effect} in {Mobile} {Sensing} {System} for {Urban} {Air} {Pollution} {Monitoring}}. In \bibinfo{booktitle}{\emph{Proceedings of the 15th {ACM} {Conference} on {Embedded} {Network} {Sensor} {Systems}}}. \bibinfo{publisher}{ACM}, \bibinfo{address}{Delft Netherlands}, \bibinfo{pages}{1--2}.
\newblock
\showISBNx{978-1-4503-5459-2}
\href{https://doi.org/10.1145/3131672.3136997}{doi:\nolinkurl{10.1145/3131672.3136997}}


\bibitem[Liu et~al\mbox{.}(2017b)]%
        {liu_individualized_2017}
\bibfield{author}{\bibinfo{person}{Xinyu Liu}, \bibinfo{person}{Xiangxiang Xu}, \bibinfo{person}{Xinlei Chen}, \bibinfo{person}{Enhan Mai}, \bibinfo{person}{Hae~Young Noh}, \bibinfo{person}{Pei Zhang}, {and} \bibinfo{person}{Lin Zhang}.} \bibinfo{year}{2017}\natexlab{b}.
\newblock \showarticletitle{Individualized {Calibration} of {Industrial}-{Grade} {Gas} {Sensors} in {Air} {Quality} {Sensing} {System}}. In \bibinfo{booktitle}{\emph{Proceedings of the 15th {ACM} {Conference} on {Embedded} {Network} {Sensor} {Systems}}}. \bibinfo{publisher}{ACM}, \bibinfo{address}{Delft Netherlands}, \bibinfo{pages}{1--2}.
\newblock
\showISBNx{978-1-4503-5459-2}
\href{https://doi.org/10.1145/3131672.3136998}{doi:\nolinkurl{10.1145/3131672.3136998}}


\bibitem[Liu et~al\mbox{.}(2024)]%
        {liu2024mobiair}
\bibfield{author}{\bibinfo{person}{Yuxuan Liu}, \bibinfo{person}{Haoyang Wang}, \bibinfo{person}{Fanhang Man}, \bibinfo{person}{Jingao Xu}, \bibinfo{person}{Fan Dang}, \bibinfo{person}{Yunhao Liu}, \bibinfo{person}{Xiao-Ping Zhang}, {and} \bibinfo{person}{Xinlei Chen}.} \bibinfo{year}{2024}\natexlab{}.
\newblock \showarticletitle{MobiAir: Unleashing Sensor Mobility for City-scale and Fine-grained Air-Quality Monitoring with AirBERT}. In \bibinfo{booktitle}{\emph{Proceedings of the 22nd Annual International Conference on Mobile Systems, Applications and Services}}. \bibinfo{pages}{223--236}.
\newblock


\bibitem[Ma et~al\mbox{.}(2019)]%
        {ma2019deep}
\bibfield{author}{\bibinfo{person}{Rui Ma}, \bibinfo{person}{Ning Liu}, \bibinfo{person}{Xiangxiang Xu}, \bibinfo{person}{Yue Wang}, \bibinfo{person}{Hae~Young Noh}, \bibinfo{person}{Pei Zhang}, {and} \bibinfo{person}{Lin Zhang}.} \bibinfo{year}{2019}\natexlab{}.
\newblock \showarticletitle{A deep autoencoder model for pollution map recovery with mobile sensing networks}. In \bibinfo{booktitle}{\emph{Adjunct Proceedings of the 2019 ACM international joint conference on pervasive and ubiquitous computing and proceedings of the 2019 ACM international symposium on wearable computers}}. \bibinfo{pages}{577--583}.
\newblock


\bibitem[Ma et~al\mbox{.}(2020a)]%
        {ma_enhancing_2020}
\bibfield{author}{\bibinfo{person}{Rui Ma}, \bibinfo{person}{Ning Liu}, \bibinfo{person}{Xiangxiang Xu}, \bibinfo{person}{Yue Wang}, \bibinfo{person}{Hae~Young Noh}, \bibinfo{person}{Pei Zhang}, {and} \bibinfo{person}{Lin Zhang}.} \bibinfo{year}{2020}\natexlab{a}.
\newblock \showarticletitle{Enhancing the {Data} {Learning} {With} {Physical} {Knowledge} in {Fine}-{Grained} {Air} {Pollution} {Inference}}.
\newblock \bibinfo{journal}{\emph{IEEE Access}}  \bibinfo{volume}{8} (\bibinfo{year}{2020}), \bibinfo{pages}{88372--88384}.
\newblock
\showISSN{2169-3536}
\href{https://doi.org/10.1109/ACCESS.2020.2993610}{doi:\nolinkurl{10.1109/ACCESS.2020.2993610}}
\newblock
\shownote{Conference Name: IEEE Access}.


\bibitem[Ma et~al\mbox{.}(2020b)]%
        {ma_fine-grained_2020}
\bibfield{author}{\bibinfo{person}{Rui Ma}, \bibinfo{person}{Ning Liu}, \bibinfo{person}{Xiangxiang Xu}, \bibinfo{person}{Yue Wang}, \bibinfo{person}{Hae~Young Noh}, \bibinfo{person}{Pei Zhang}, {and} \bibinfo{person}{Lin Zhang}.} \bibinfo{year}{2020}\natexlab{b}.
\newblock \showarticletitle{Fine-{Grained} {Air} {Pollution} {Inference} with {Mobile} {Sensing} {Systems}: {A} {Weather}-{Related} {Deep} {Autoencoder} {Model}}.
\newblock \bibinfo{journal}{\emph{Proc. ACM Interact. Mob. Wearable Ubiquitous Technol.}} \bibinfo{volume}{4}, \bibinfo{number}{2} (\bibinfo{date}{June} \bibinfo{year}{2020}), \bibinfo{pages}{52:1--52:21}.
\newblock
\href{https://doi.org/10.1145/3397322}{doi:\nolinkurl{10.1145/3397322}}


\bibitem[Maag et~al\mbox{.}(2018)]%
        {maag_survey_2018}
\bibfield{author}{\bibinfo{person}{Balz Maag}, \bibinfo{person}{Zimu Zhou}, {and} \bibinfo{person}{Lothar Thiele}.} \bibinfo{year}{2018}\natexlab{}.
\newblock \showarticletitle{A Survey on Sensor Calibration in Air Pollution Monitoring Deployments}.
\newblock  \bibinfo{volume}{5}, \bibinfo{number}{6} (\bibinfo{year}{2018}), \bibinfo{pages}{4857--4870}.
\newblock
\showISSN{2327-4662}
\href{https://doi.org/10.1109/JIOT.2018.2853660}{doi:\nolinkurl{10.1109/JIOT.2018.2853660}}


\bibitem[Maag et~al\mbox{.}(2018)]%
        {maag_w-air_2018}
\bibfield{author}{\bibinfo{person}{Balz Maag}, \bibinfo{person}{Zimu Zhou}, {and} \bibinfo{person}{Lothar Thiele}.} \bibinfo{year}{2018}\natexlab{}.
\newblock \showarticletitle{W-{Air}: {Enabling} {Personal} {Air} {Pollution} {Monitoring} on {Wearables}}.
\newblock \bibinfo{journal}{\emph{Proc. ACM Interact. Mob. Wearable Ubiquitous Technol.}} \bibinfo{volume}{2}, \bibinfo{number}{1} (\bibinfo{date}{March} \bibinfo{year}{2018}), \bibinfo{pages}{24:1--24:25}.
\newblock
\href{https://doi.org/10.1145/3191756}{doi:\nolinkurl{10.1145/3191756}}


\bibitem[M{\'e}ndez et~al\mbox{.}(2023)]%
        {mendez2023machine}
\bibfield{author}{\bibinfo{person}{Manuel M{\'e}ndez}, \bibinfo{person}{Mercedes~G Merayo}, {and} \bibinfo{person}{Manuel N{\'u}{\~n}ez}.} \bibinfo{year}{2023}\natexlab{}.
\newblock \showarticletitle{Machine learning algorithms to forecast air quality: a survey}.
\newblock \bibinfo{journal}{\emph{Artificial Intelligence Review}} \bibinfo{volume}{56}, \bibinfo{number}{9} (\bibinfo{year}{2023}), \bibinfo{pages}{10031--10066}.
\newblock


\bibitem[Moltchanov et~al\mbox{.}(2015)]%
        {moltchanov_feasibility_2015}
\bibfield{author}{\bibinfo{person}{Sharon Moltchanov}, \bibinfo{person}{Ilan Levy}, \bibinfo{person}{Yael Etzion}, \bibinfo{person}{Uri Lerner}, \bibinfo{person}{David~M. Broday}, {and} \bibinfo{person}{Barak Fishbain}.} \bibinfo{year}{2015}\natexlab{}.
\newblock \showarticletitle{On the feasibility of measuring urban air pollution by wireless distributed sensor networks}.
\newblock \bibinfo{journal}{\emph{Science of The Total Environment}}  \bibinfo{volume}{502} (\bibinfo{date}{Jan.} \bibinfo{year}{2015}), \bibinfo{pages}{537--547}.
\newblock
\showISSN{0048-9697}
\href{https://doi.org/10.1016/j.scitotenv.2014.09.059}{doi:\nolinkurl{10.1016/j.scitotenv.2014.09.059}}


\bibitem[Moore et~al\mbox{.}(2012)]%
        {moore_air_2012}
\bibfield{author}{\bibinfo{person}{Adam Moore}, \bibinfo{person}{Miguel Figliozzi}, {and} \bibinfo{person}{Christopher~M. Monsere}.} \bibinfo{year}{2012}\natexlab{}.
\newblock \showarticletitle{Air {Quality} at {Bus} {Stops}: {Empirical} {Analysis} of {Exposure} to {Particulate} {Matter} at {Bus} {Stop} {Shelters}}.
\newblock \bibinfo{journal}{\emph{Transportation Research Record}} \bibinfo{volume}{2270}, \bibinfo{number}{1} (\bibinfo{date}{Jan.} \bibinfo{year}{2012}), \bibinfo{pages}{76--86}.
\newblock
\showISSN{0361-1981}
\href{https://doi.org/10.3141/2270-10}{doi:\nolinkurl{10.3141/2270-10}}
\newblock
\shownote{Publisher: SAGE Publications Inc}.


\bibitem[Munera et~al\mbox{.}(2021)]%
        {munera2021iot}
\bibfield{author}{\bibinfo{person}{Danny Munera}, \bibinfo{person}{Johnny Aguirre}, \bibinfo{person}{Natalia~Gaviria Gomez}, {et~al\mbox{.}}} \bibinfo{year}{2021}\natexlab{}.
\newblock \showarticletitle{IoT-based air quality monitoring systems for smart cities: A systematic mapping study}.
\newblock \bibinfo{journal}{\emph{International Journal of Electrical and Computer Engineering}} \bibinfo{volume}{11}, \bibinfo{number}{4} (\bibinfo{year}{2021}), \bibinfo{pages}{3470}.
\newblock


\bibitem[Narayana et~al\mbox{.}(2023)]%
        {narayana_sens-bert_2023}
\bibfield{author}{\bibinfo{person}{M.~V. Narayana}, \bibinfo{person}{Kranthi~Kumar Rachvarapu}, \bibinfo{person}{Devendra Jalihal}, {and} \bibinfo{person}{Shiva Nagendra~S. M}.} \bibinfo{year}{2023}\natexlab{}.
\newblock \bibinfo{title}{Sens-{BERT}: Enabling Transferability and Re-calibration of Calibration Models for Low-cost Sensors under Reference Measurements Scarcity}.
\newblock
\href{https://doi.org/10.48550/arXiv.2309.13390}{doi:\nolinkurl{10.48550/arXiv.2309.13390}}
\showeprint[arxiv]{2309.13390}


\bibitem[Nikzad et~al\mbox{.}(2012)]%
        {nikzad_citisense_2012}
\bibfield{author}{\bibinfo{person}{Nima Nikzad}, \bibinfo{person}{Nakul Verma}, \bibinfo{person}{Celal Ziftci}, \bibinfo{person}{Elizabeth Bales}, \bibinfo{person}{Nichole Quick}, \bibinfo{person}{Piero Zappi}, \bibinfo{person}{Kevin Patrick}, \bibinfo{person}{Sanjoy Dasgupta}, \bibinfo{person}{Ingolf Krueger}, \bibinfo{person}{Tajana~Šimunić Rosing}, {and} \bibinfo{person}{William~G. Griswold}.} \bibinfo{year}{2012}\natexlab{}.
\newblock \showarticletitle{{CitiSense}: improving geospatial environmental assessment of air quality using a wireless personal exposure monitoring system}. In \bibinfo{booktitle}{\emph{Proceedings of the conference on {Wireless} {Health}}} \emph{(\bibinfo{series}{{WH} '12})}. \bibinfo{publisher}{Association for Computing Machinery}, \bibinfo{address}{New York, NY, USA}, \bibinfo{pages}{1--8}.
\newblock
\showISBNx{978-1-4503-1760-3}
\href{https://doi.org/10.1145/2448096.2448107}{doi:\nolinkurl{10.1145/2448096.2448107}}


\bibitem[Noor et~al\mbox{.}(2024)]%
        {noor2024fusion}
\bibfield{author}{\bibinfo{person}{Alam Noor}, \bibinfo{person}{Kai Li}, \bibinfo{person}{Eduardo Tovar}, \bibinfo{person}{Pei Zhang}, {and} \bibinfo{person}{Bo Wei}.} \bibinfo{year}{2024}\natexlab{}.
\newblock \showarticletitle{Fusion flow-enhanced graph pooling residual networks for unmanned aerial vehicles surveillance in day and night dual visions}.
\newblock \bibinfo{journal}{\emph{Engineering Applications of Artificial Intelligence}}  \bibinfo{volume}{136} (\bibinfo{year}{2024}), \bibinfo{pages}{108959}.
\newblock


\bibitem[Shirai et~al\mbox{.}(2016)]%
        {shirai_toward_2016}
\bibfield{author}{\bibinfo{person}{Yoshinari Shirai}, \bibinfo{person}{Yasue Kishino}, \bibinfo{person}{Futoshi Naya}, {and} \bibinfo{person}{Yutaka Yanagisawa}.} \bibinfo{year}{2016}\natexlab{}.
\newblock \showarticletitle{Toward {On}-{Demand} {Urban} {Air} {Quality} {Monitoring} using {Public} {Vehicles}}. In \bibinfo{booktitle}{\emph{Proceedings of the 2nd {International} {Workshop} on {Smart}}} \emph{(\bibinfo{series}{{SmartCities} '16})}. \bibinfo{publisher}{Association for Computing Machinery}, \bibinfo{address}{New York, NY, USA}, \bibinfo{pages}{1--6}.
\newblock
\showISBNx{978-1-4503-4667-2}
\href{https://doi.org/10.1145/3009912.3009920}{doi:\nolinkurl{10.1145/3009912.3009920}}


\bibitem[Si et~al\mbox{.}(2020)]%
        {si_evaluation_2020}
\bibfield{author}{\bibinfo{person}{Minxing Si}, \bibinfo{person}{Ying Xiong}, \bibinfo{person}{Shan Du}, {and} \bibinfo{person}{Ke Du}.} \bibinfo{year}{2020}\natexlab{}.
\newblock \showarticletitle{Evaluation and calibration of a low-cost particle sensor in ambient conditions using machine-learning methods}.
\newblock  \bibinfo{volume}{13}, \bibinfo{number}{4} (\bibinfo{year}{2020}), \bibinfo{pages}{1693--1707}.
\newblock
\showISSN{1867-8548}
\href{https://doi.org/10.5194/amt-13-1693-2020}{doi:\nolinkurl{10.5194/amt-13-1693-2020}}


\bibitem[Song et~al\mbox{.}(2023)]%
        {song_spatial_2023}
\bibfield{author}{\bibinfo{person}{Yingqiang Song}, \bibinfo{person}{Changjian Zhang}, \bibinfo{person}{Xin Jin}, \bibinfo{person}{Xiaoyu Zhao}, \bibinfo{person}{Wei Huang}, \bibinfo{person}{Xiaoshuang Sun}, \bibinfo{person}{Zhongkang Yang}, {and} \bibinfo{person}{Shuhuan Wang}.} \bibinfo{year}{2023}\natexlab{}.
\newblock \showarticletitle{Spatial prediction of {PM}2.5 concentration using hyper-parameter optimization {XGBoost} model in China}.
\newblock   \bibinfo{volume}{32} (\bibinfo{year}{2023}), \bibinfo{pages}{103272}.
\newblock
\showISSN{2352-1864}
\href{https://doi.org/10.1016/j.eti.2023.103272}{doi:\nolinkurl{10.1016/j.eti.2023.103272}}


\bibitem[Tarwidi et~al\mbox{.}(2023)]%
        {tarwidi_optimized_2023}
\bibfield{author}{\bibinfo{person}{Dede Tarwidi}, \bibinfo{person}{Sri~Redjeki Pudjaprasetya}, \bibinfo{person}{Didit Adytia}, {and} \bibinfo{person}{Mochamad Apri}.} \bibinfo{year}{2023}\natexlab{}.
\newblock \showarticletitle{An optimized {XGBoost}-based machine learning method for predicting wave run-up on a sloping beach}.
\newblock   \bibinfo{volume}{10} (\bibinfo{year}{2023}), \bibinfo{pages}{102119}.
\newblock
\showISSN{22150161}
\href{https://doi.org/10.1016/j.mex.2023.102119}{doi:\nolinkurl{10.1016/j.mex.2023.102119}}


\bibitem[Tsujita et~al\mbox{.}(2005)]%
        {tsujita_gas_2005}
\bibfield{author}{\bibinfo{person}{Wataru Tsujita}, \bibinfo{person}{Akihito Yoshino}, \bibinfo{person}{Hiroshi Ishida}, {and} \bibinfo{person}{Toyosaka Moriizumi}.} \bibinfo{year}{2005}\natexlab{}.
\newblock \showarticletitle{Gas sensor network for air-pollution monitoring}.
\newblock \bibinfo{journal}{\emph{Sensors and Actuators B: Chemical}} \bibinfo{volume}{110}, \bibinfo{number}{2} (\bibinfo{date}{Oct.} \bibinfo{year}{2005}), \bibinfo{pages}{304--311}.
\newblock
\showISSN{0925-4005}
\href{https://doi.org/10.1016/j.snb.2005.02.008}{doi:\nolinkurl{10.1016/j.snb.2005.02.008}}


\bibitem[Tunca et~al\mbox{.}(2023)]%
        {tunca_calibrating_2023}
\bibfield{author}{\bibinfo{person}{Emre Tunca}, \bibinfo{person}{Eyüp~Selim Köksal}, {and} \bibinfo{person}{Sakine Çetin Taner}.} \bibinfo{year}{2023}\natexlab{}.
\newblock \showarticletitle{Calibrating {UAV} thermal sensors using machine learning methods for improved accuracy in agricultural applications}.
\newblock   \bibinfo{volume}{133} (\bibinfo{year}{2023}), \bibinfo{pages}{104804}.
\newblock
\showISSN{1350-4495}
\href{https://doi.org/10.1016/j.infrared.2023.104804}{doi:\nolinkurl{10.1016/j.infrared.2023.104804}}


\bibitem[Van~Poppel et~al\mbox{.}(2023)]%
        {van_poppel_senseurcity_2023}
\bibfield{author}{\bibinfo{person}{Martine Van~Poppel}, \bibinfo{person}{Philipp Schneider}, \bibinfo{person}{Jan Peters}, \bibinfo{person}{Sinan Yatkin}, \bibinfo{person}{Michel Gerboles}, \bibinfo{person}{Christina Matheeussen}, \bibinfo{person}{Alena Bartonova}, \bibinfo{person}{Silvije Davila}, \bibinfo{person}{Marco Signorini}, \bibinfo{person}{Matthias Vogt}, \bibinfo{person}{Franck~René Dauge}, \bibinfo{person}{Jøran~Solnes Skaar}, {and} \bibinfo{person}{Rolf Haugen}.} \bibinfo{year}{2023}\natexlab{}.
\newblock \showarticletitle{{SensEURCity}: A multi-city air quality dataset collected for 2020/2021 using open low-cost sensor systems}.
\newblock  \bibinfo{volume}{10}, \bibinfo{number}{1} (\bibinfo{year}{2023}), \bibinfo{pages}{322}.
\newblock
\showISSN{2052-4463}
\href{https://doi.org/10.1038/s41597-023-02135-w}{doi:\nolinkurl{10.1038/s41597-023-02135-w}}


\bibitem[Villanueva et~al\mbox{.}(2023)]%
        {villanueva_smart_2023}
\bibfield{author}{\bibinfo{person}{Edwin Villanueva}, \bibinfo{person}{Soledad Espezua}, \bibinfo{person}{George Castelar}, \bibinfo{person}{Kyara Diaz}, {and} \bibinfo{person}{Erick Ingaroca}.} \bibinfo{year}{2023}\natexlab{}.
\newblock \showarticletitle{Smart Multi-Sensor Calibration of Low-Cost Particulate Matter Monitors}.
\newblock  \bibinfo{volume}{23}, \bibinfo{number}{7} (\bibinfo{year}{2023}), \bibinfo{pages}{3776}.
\newblock
\showISSN{1424-8220}
\href{https://doi.org/10.3390/s23073776}{doi:\nolinkurl{10.3390/s23073776}}


\bibitem[{World Health Organization (WHO)}(nd)]%
        {who_air_pollution}
\bibfield{author}{\bibinfo{person}{{World Health Organization (WHO)}}.} \bibinfo{year}{n.d.}\natexlab{}.
\newblock \bibinfo{title}{Air Pollution Data Portal}.
\newblock \bibinfo{howpublished}{\url{https://www.who.int/data/gho/data/themes/air-pollution}}.
\newblock
\newblock
\shownote{Accessed: 2025-03-29}.


\bibitem[Wu et~al\mbox{.}(2020)]%
        {wu_generative_2020}
\bibfield{author}{\bibinfo{person}{Zhengwei Wu}, \bibinfo{person}{Xiaoxi Zhang}, \bibinfo{person}{Susu Xu}, \bibinfo{person}{Xinlei Chen}, \bibinfo{person}{Pei Zhang}, \bibinfo{person}{Hae~Young Noh}, {and} \bibinfo{person}{Carlee Joe-Wong}.} \bibinfo{year}{2020}\natexlab{}.
\newblock \showarticletitle{A generative simulation platform for multi-agent systems with incentives}. In \bibinfo{booktitle}{\emph{Adjunct {Proceedings} of the 2020 {ACM} {International} {Joint} {Conference} on {Pervasive} and {Ubiquitous} {Computing} and {Proceedings} of the 2020 {ACM} {International} {Symposium} on {Wearable} {Computers}}}. \bibinfo{publisher}{ACM}, \bibinfo{address}{Virtual Event Mexico}, \bibinfo{pages}{580--587}.
\newblock
\showISBNx{9781450380768}
\href{https://doi.org/10.1145/3410530.3414590}{doi:\nolinkurl{10.1145/3410530.3414590}}


\bibitem[Xu et~al\mbox{.}(2019a)]%
        {xu_incentivizing_2019}
\bibfield{author}{\bibinfo{person}{Susu Xu}, \bibinfo{person}{Xinlei Chen}, \bibinfo{person}{Xidong Pi}, \bibinfo{person}{Carlee Joe-Wong}, \bibinfo{person}{Pei Zhang}, {and} \bibinfo{person}{Hae~Young Noh}.} \bibinfo{year}{2019}\natexlab{a}.
\newblock \showarticletitle{Incentivizing vehicular crowdsensing system for large scale smart city applications}. In \bibinfo{booktitle}{\emph{Sensors and {Smart} {Structures} {Technologies} for {Civil}, {Mechanical}, and {Aerospace} {Systems} 2019}}, Vol.~\bibinfo{volume}{10970}. \bibinfo{publisher}{SPIE}, \bibinfo{pages}{402--408}.
\newblock
\href{https://doi.org/10.1117/12.2514021}{doi:\nolinkurl{10.1117/12.2514021}}


\bibitem[Xu et~al\mbox{.}(2019b)]%
        {xu_vehicle_2019}
\bibfield{author}{\bibinfo{person}{Susu Xu}, \bibinfo{person}{Xinlei Chen}, \bibinfo{person}{Xidong Pi}, \bibinfo{person}{Carlee Joe-Wong}, \bibinfo{person}{Pei Zhang}, {and} \bibinfo{person}{Hae~Young Noh}.} \bibinfo{year}{2019}\natexlab{b}.
\newblock \showarticletitle{Vehicle dispatching for sensing coverage optimization in mobile crowdsensing systems: poster abstract}. In \bibinfo{booktitle}{\emph{Proceedings of the 18th {International} {Conference} on {Information} {Processing} in {Sensor} {Networks}}}. \bibinfo{publisher}{ACM}, \bibinfo{address}{Montreal Quebec Canada}, \bibinfo{pages}{311--312}.
\newblock
\showISBNx{978-1-4503-6284-9}
\href{https://doi.org/10.1145/3302506.3312604}{doi:\nolinkurl{10.1145/3302506.3312604}}


\bibitem[Xu et~al\mbox{.}(2016)]%
        {xu_gotcha_2016}
\bibfield{author}{\bibinfo{person}{Xiangxiang Xu}, \bibinfo{person}{Xinlei Chen}, \bibinfo{person}{Xinyu Liu}, \bibinfo{person}{Hae~Young Noh}, \bibinfo{person}{Pei Zhang}, {and} \bibinfo{person}{Lin Zhang}.} \bibinfo{year}{2016}\natexlab{}.
\newblock \showarticletitle{Gotcha {II}: {Deployment} of a {Vehicle}-based {Environmental} {Sensing} {System}: {Poster} {Abstract}}. In \bibinfo{booktitle}{\emph{Proceedings of the 14th {ACM} {Conference} on {Embedded} {Network} {Sensor} {Systems} {CD}-{ROM}}}. \bibinfo{publisher}{ACM}, \bibinfo{address}{Stanford CA USA}, \bibinfo{pages}{376--377}.
\newblock
\showISBNx{978-1-4503-4263-6}
\href{https://doi.org/10.1145/2994551.2996714}{doi:\nolinkurl{10.1145/2994551.2996714}}


\bibitem[Yin et~al\mbox{.}(2025)]%
        {yin2025survey}
\bibfield{author}{\bibinfo{person}{Kevin Yin}, \bibinfo{person}{Kevin Maffetone}, \bibinfo{person}{Yahya Naveed}, {and} \bibinfo{person}{Julia Gersey}.} \bibinfo{year}{2025}\natexlab{}.
\newblock \bibinfo{booktitle}{\emph{Survey of Computational and Sensing Methods for Pre-emergent Zoonotic Viral Pathogen Tracking}}.
\newblock
\urldef\tempurl%
\url{https://www.researchgate.net/publication/390027752_Survey_of_Computational_and_Sensing_Methods_for_Pre-emergent_Zoonotic_Viral_Pathogen_Tracking}
\showURL{%
\tempurl}


\end{thebibliography}

\end{document}